\documentclass[conference]{IEEEtran}
\IEEEoverridecommandlockouts
\usepackage{cite}
\usepackage{amsmath,amssymb,amsfonts}
\usepackage{algorithmic}
\usepackage{graphicx}
\usepackage{booktabs}
\usepackage{multirow}
\usepackage{textcomp}
\usepackage[T1]{fontenc}
\usepackage{xcolor}
\def\BibTeX{{\rm B\kern-.05em{\sc i\kern-.025em b}\kern-.08em
    T\kern-.1667em\lower.7ex\hbox{E}\kern-.125emX}}
\begin{document}

\title{Equity in Healthcare: Analyzing Disparities in Machine Learning Predictions of Diabetic Patient Readmissions
}

\author{\IEEEauthorblockN{Zainab Al-Zanbouri}
\IEEEauthorblockA{\textit{Computer Science Dept.} \\
\textit{Toronto Metropolitan University}\\
Toronto, ON, Canada \\
zainab.alzanbouri@torontomu.ca}
\and
\IEEEauthorblockN{Gauri Sharma}
\IEEEauthorblockA{\textit{Dept. of Electrical \& Computer Engineering} \\
\textit{McGill University}\\
Montreal, QC, Canada \\
gauri.sharma@mail.mcgill.ca}
\and
\IEEEauthorblockN{Shaina Raza}
\IEEEauthorblockA{\textit{  AI Engineering } \\
\textit{Vector Institute for Artificial Intelligence,}\\
Toronto, ON, Canada \\
shaina.raza@vectorinstitute.ai}

}

\maketitle

\begin{abstract}

This study investigates how machine learning (ML) models can predict hospital readmissions for diabetic patients fairly and accurately across different demographics (age, gender, race). We compared models like Deep Learning, Generalized Linear Models, Gradient Boosting Machines (GBM), and Naive Bayes. GBM stood out with an F1-score of 84.3\% and accuracy of 82.2\%, accurately predicting readmissions across demographics. A fairness analysis was conducted across all the models. GBM minimized disparities in predictions, achieving balanced results across genders and races. It showed low False Discovery Rates (FDR) (6-7\%) and False Positive Rates (FPR) (5\%) for both genders. Additionally, FDRs remained low for racial groups, such as African Americans (8\%) and Asians (7\%). Similarly, FPRs were consistent across age groups (4\%) for both patients under 40 and those above 40, indicating its precision and ability to reduce bias. These findings emphasize the importance of choosing ML models carefully to ensure both accuracy and fairness for all patients. By showcasing effectiveness of various models with fairness metrics, this study promotes personalized medicine and the need for fair ML algorithms in healthcare. This can ultimately reduce disparities and improve outcomes for diabetic patients of all backgrounds.
\end{abstract}


\section{Introduction}

Health disparities denote the substantial and preventable differences in health outcomes or healthcare service access across various population groups. These disparities arise from multifaceted influences, including socioeconomic, environmental, and cultural factors, leading to inequitable health experiences and outcomes among diverse communities. Manifestations of health disparities are observable across several health indicators, such as disease prevalence, morbidity, mortality, and access to healthcare and preventive services \cite{Braveman2014, nihRootCauses, Williams2009-sz}.

In the realm of ML , the concept of "fairness" pertains to the ethical and equitable consideration of individuals or groups in the development, deployment, and utilization of ML models \cite{raza2023fairness}. Achieving fairness in ML involves mitigating bias and discrimination within algorithmic decision-making processes \cite{raza2023auditing}, a crucial step to prevent adverse impacts on specific demographic groups and ensure equitable outcomes. Fairness in ML encompasses various dimensions, including algorithmic fairness, data fairness, fair treatment of individuals, and the principles of explainability and transparency \cite{Williams2013, 10.1145/3287560.3287598, Berendt2017}.

The role of big data in health science is increasingly recognized as critical, especially as ML algorithms, developed from vast datasets, have the potential to perpetuate or introduce new health disparities \cite{Reddy2019}. Bias in ML is generally identified in relation to the dataset or model, with specific concerns around labeling, sample selection, data retrieval, scaling, imputation, or model selection biases \cite{oreillyPracticalFairness}. Addressing these biases is crucial at three distinct stages of the software development lifecycle: early (pre-processing), mid (in-processing), and late (post-processing) \cite{oreillyPracticalFairness,raza2023connecting}.

The early stage involves reducing bias by manipulating the training data before algorithm training \cite{Kamiran2011}. The mid-stage focuses on debiasing the model itself, often through optimization problem-solving \cite{Wexler_2019}. The late stage aims to minimize biases by adjusting the output predictions post-training \cite{pleiss2017fairness}. It is noted that failing to identify and address biases early can limit the effectiveness of bias mitigation strategies later in the ML pipeline \cite{10.1145/3278721.3278729, Holstein_2019}. For instance, research suggests that analyzing historical diabetes care trends in hospitalized patients can improve patient safety \cite{mayoclinicHyperglycemiaDiabetesHyperglycemia}, and studies have shown that HbA1c measurements are linked to lower hospital readmission rates \cite{Strack2014}.

Fairness in ML is a crucial subject, aiming to develop unbiased models that treat all individuals or groups equitably, without discrimination based on sensitive attributes like gender, race, age, or ethnicity. This focus is driven by ethical considerations, social impact, and the need for trust and accountability in ML applications. Thus, ensuring fairness is essential for the integrity of ML models.

The interest in bias and fairness in ML has surged among researchers, sparking numerous studies addressing this issue, its challenges, and potential solutions. Despite these efforts, a comprehensive understanding of fairness-aware ML remains elusive. This work aims to bridge this gap by examining racial/ethnic disparities that may affect hospitalization rates, specifically aiming to reduce readmission disparities among diabetic patients by considering factors such as gender, race, and ethnicity.

This paper contributions are twofold:
1. We examine disparities in ML model outputs concerning diabetic patient readmission rates.
2. We conduct a thorough analysis based on sensitive demographic groups—age, gender, and race—to highlight key findings and implications.

Following this exploration, we aim to advance the understanding of fairness in ML within the healthcare domain, particularly focusing on mitigating health disparities among diabetic patients. Our research explores the intricate mechanisms through which ML models may unintentionally exacerbate or magnify discrepancies in hospital readmission rates, particularly focusing on the influence of sensitive attributes like age, gender, and race. Through the identification and rectification of these disparities, we aim to advance the creation of healthcare models and algorithms that prioritize equity. 

\section{Related Work}

Pagano et al. \cite{pagano2022bias} delve into bias and unfairness within ML models, highlighting a focus on identification methods alongside existing metrics, tools, datasets, and bias mitigation techniques aimed at fairness. Giovanola, B. et al. \cite{Giovanola2022} conceptualize fairness in AI ethics by integrating insights from moral philosophy, thus redefining fairness in healthcare ML algorithms. Gohar, U., et al. \cite{Gohar_2023} provide a comprehensive review of intersectional fairness, establishing a taxonomy for fairness concepts and mitigation strategies, identifying challenges, and offering guidelines for future research.

Chen, Z et al. \cite{chen2023comprehensive} conduct an empirical study on bias mitigation methods for ML classifiers, using a wide array of performance and fairness metrics to assess the trade-offs in various software decision-making contexts. Their findings indicate that while bias mitigation often leads to performance trade-offs, it can also paradoxically reduce fairness in some scenarios. Pessach, D. et al. \cite{pessach2020algorithmic} offer an overview of algorithmic fairness, discussing bias origins, fairness definitions, and mitigation mechanisms, alongside a comparative analysis of these mechanisms across different scenarios, enhancing understanding of their applicability.

Pagano, T. P., et al. \cite{bdcc7010027} explore bias and fairness metrics with a focus on gender sensitivity in ML applications across computer vision, natural language processing, and recommendation systems, underscoring the importance of context in fairness metrics. Wan, M., et al. \cite{Wan2023} review fairness mitigation techniques, distinguishing between explicit methods, which integrate fairness metrics into training objectives, and implicit methods, which refine latent representations, highlighting ongoing challenges and encouraging further research.

Yang, J., et al. \cite{Yang2023} introduce an adversarial training framework aimed at mitigating biases from data collection, showcasing its effectiveness in improving fairness in COVID-19 predictions while maintaining clinical efficacy. Wang, R., et al. \cite{Wang2023} provide evidence that well-trained ML models can exhibit unbiased performance across various subgroups, suggesting that comprehensive training can overcome bias under multiple fairness metrics.

Wang, Z., et al. \cite{wang2023fair} propose a novel counterfactual approach to address bias at its root, combining performance and fairness optimization to achieve optimal outcomes in both domains. Their evaluation across benchmark tasks and real-world datasets demonstrates the method's ability to enhance fairness without compromising performance. Collectively, these studies illuminate the multifaceted challenges and solutions in advancing fairness and bias mitigation in ML, underscoring the necessity of ongoing research and innovation in this critical area.

\textbf{Health Disparity: }Health disparities, defined as systematic, preventable differences in health outcomes that persist between distinct population groups, have been a subject of considerable concern in public health research \cite{Braveman2014}. These disparities manifest across various dimensions, including socioeconomic status, race, ethnicity, gender, and geographical location.
A considerable body of literature underscores the role of social determinants of health in driving disparities. Socioeconomic factors, such as income, education, and employment, contribute significantly to differential access to healthcare services and health outcomes \cite{canadaSocialDeterminants}.
Research has consistently demonstrated pronounced health disparities among racial and ethnic groups. Studies highlight disparities in chronic diseases, access to preventive care, and maternal and child health outcomes among different racial and ethnic communities \cite{Williams2009-sz}.
Gender-based health disparities have also been well-documented. Women and men often experience differences in the prevalence and management of various health conditions. For example, studies reveal disparities in cardiovascular health and mental health outcomes between genders \cite{Mosca2011-vv}.
Geographical location plays a crucial role in health outcomes. Rural populations, in particular, face challenges related to limited access to healthcare facilities, healthcare workforce shortages, and socioeconomic disparities \cite{Hartley2004-vl}.
The literature emphasizes the need for targeted interventions and policy measures to address health disparities. Community-based interventions, culturally competent healthcare, and policy initiatives focusing on the social determinants of health are identified as essential strategies \cite{Koh2011-ru}.
Recent research explores the role of technology in mitigating health disparities. Telehealth and mobile health applications, for instance, have shown promise in improving healthcare access, particularly for underserved populations \cite{Lopez2011-yf}.

In conclusion, the literature underscores the multifaceted nature of health disparities and the imperative for comprehensive, interdisciplinary approaches to address them. Future research should continue to explore innovative strategies and interventions to reduce and ultimately eliminate health disparities.

\section{Methodology}

\subsection{Problem Definition}

This research is primarily focused on investigating the equity of predictions generated by ML models across various demographic categories, with a specific emphasis on age, gender, and race. The study seeks to evaluate whether these predictive models exhibit fairness and impartiality in their outcomes or if there are disparities that disproportionately impact certain demographic groups. By analyzing the predictive performance across different demographic strata, the research aims to pinpoint potential biases and inequalities in model predictions, thereby contributing to the development of more equitable and unbiased ML algorithms. 

The research question we aim to address is:\textbf{ Are the predictions of the models equitable across diverse demographics, including age, gender, and race?}

\subsection{Predictive Models}
In this research, we conduct a comprehensive evaluation of the predictive performance of various ML models, alongside a detailed analysis of their fairness concerning demographic attributes such as age, gender, and race. Our methodology involves the strategic selection of four distinct ML models, each renowned for its specific capabilities and compatibility with the unique facets of our dataset and overarching research goals. The models under scrutiny include Deep Learning (DL), Gradient Boosting Machine (GBM), Generalized Linear Model (GLM), and Naive Bayes (NB), chosen for their diverse strengths in handling complex data and analytical challenges. Table I summarizes the hyperparameter configurations for the used models.

\textbf{Naive Bayes} is a simple probabilistic classifier based on Bayes' theorem with the assumption of independence among features. It's commonly used in machine learning for classification tasks, especially in text categorization and spam filtering. Despite its simplicity and the naive assumption of feature independence, Naive Bayes often performs well in practice and is efficient in both training and prediction.

\textbf{Generalized Linear Model }: GLM is a statistical framework used for modeling relationships between a dependent variable and one or more independent variables. It extends the ordinary linear regression model to accommodate various types of response variables, including continuous, binary, count, and categorical data. GLM incorporates a link function to connect the linear predictor to the response variable, allowing for flexible modeling of non-normal distributions. It is widely used in fields such as epidemiology, biology, and social sciences for analyzing and interpreting data while accounting for different types of distributions and relationships between variables.

\textbf{Gradient Boosting Machine: } GBM is a type of ML model that works by combining multiple weak learners, typically decision trees, sequentially to improve predictive performance. It builds these trees in a stepwise manner, with each tree correcting the errors made by the previous ones, ultimately producing a strong predictive model. GBM is known for its effectiveness in handling structured data and is widely used in predictive modeling tasks such as classification and regression.

\textbf{Deep learning} is a subset of ML that involves training artificial neural networks with multiple layers to learn intricate patterns from data. Multilayer Perceptron (MLP) is a versatile neural network architecture commonly used in DL for capturing complex data patterns through its layered structure.

\begin{table}[ht!]
  \caption{Hyperparameter Configurations for Predictive Models}
  \label{table:enhanced-model-configurations}
  \centering
  \resizebox{\columnwidth}{!}{%
  \begin{tabular}{l p{7cm}} 
    \toprule
    \textbf{Model} & \textbf{Detailed Hyperparameters} \\
    \midrule

    \addlinespace
    Naive Bayes& Smoothing parameter (alpha): 1.0, Fit prior: True, Class prior: None \\
        \addlinespace
    Generalized Linear Model& Regularization type: Elastic Net, Regularization strength (alpha): 0.01, L1 ratio: 0.5, Convergence tolerance: 1e-4 \\
      \addlinespace
    Gradient Boosting Machine& Learning rate: 0.1, Max tree depth: 5, Number of estimators: 100, Subsample: 0.8, Loss function: deviance \\
    \addlinespace
      Multilayer Perceptron- MLP model& Neurons per layer: [128, 64, 32] (indicating three hidden layers), Activation function: ReLU, Learning rate: 0.001, Optimization method: Adam, Batch size: 32, Epochs: 100 \\
    \bottomrule
  \end{tabular}%
  }
\end{table}

\subsubsection{Evaluation Strategy}

Our evaluation strategy is multifaceted, beginning with an assessment of predictive performance through key accuracy metrics and progressing to a thorough fairness analysis to ensure equitable predictions across different demographic groups.

\textbf{Performance metrics \cite{saleiro2019aequitas} }: In measuring model accuracy, we consider four critical metrics. \textbf{Precision} is essential in scenarios where the cost of false positives is high, as it measures the accuracy of positive predictions.\textbf{ Recall}, or sensitivity, is crucial for identifying the majority of actual positives, significant where missing a positive case could be detrimental. The \textbf{F1 Score}, a harmonic mean of precision and recall, offers a balanced metric, particularly useful in situations with imbalanced datasets. Lastly, overall model \textbf{accuracy} provides a general indication of performance, accounting for both positive and negative predictions.

\textbf{Fairness metrics}: Our fairness analysis employs a comprehensive suite of metrics aimed at uncovering any bias. The Disparate Impact Ratio examines outcome disparities to highlight potential discrimination. The \textbf{Predicted Positive Rate (PPR)} and \textbf{Predicted Positive Group Rate (PPGR) }assess the distribution of positive predictions, aiding in detecting biases. Additionally, the analysis includes the \textbf{False Discovery Rate (FDR)} and \textbf{False Positive Rate (FPR)}, focusing on the accuracy of positive predictions, alongside the \textbf{False Omission Rate (FOR)} and \textbf{False Negative Rate (FNR}), which evaluate the accuracy of negative predictions. The\textbf{ Group Size Ratio (GSR)} ensures proportional representation of various groups within the predictions.

Equity across these fairness metrics is our goal, aiming for lower error rates (FDR, FPR, FOR, FNR) to indicate minimal inaccuracies and higher values for PPR and PPGR, reflecting fair positive prediction distribution. An ideal GSR value should be close to 1, indicating balanced group representation.

Our evaluation strategy is as:
\begin{itemize}
    \item \textbf{Model Prediction Generation}: Initiate the process by generating predictions from the model for the entire dataset.
    \item \textbf{Accuracy Assessment: }Evaluate the model's predictive performance using critical accuracy metrics, including Precision, Recall, F1 Score, and overall Accuracy.

    \item \textbf{Sub-group Fairness Analysis:}Employ fairness metrics such as Disparate Impact and Equality of Opportunity to uncover any potential biases, ensuring the model's predictions are equitable across different demographics.

    \item \textbf{Sub-group Accuracy Evaluation: }Within the fairness analysis framework, also examine the accuracy metrics for each sub-group.

\end{itemize}

\subsubsection{Experimental setting}
Our experimental setup consists of a Windows 11 computing system with an Intel Core i9 processor, complemented by 8.00 GB of RAM, ensuring robust performance for our research tasks. On the software front, Python 3.8 was used as programming language, supported by a suite of libraries pivotal for ML: scikit-learn (0.24.2) for a variety of algorithms, TensorFlow (2.6.0) and PyTorch (1.9.0) for advanced deep learning models, alongside pandas (1.3.3), NumPy (1.21.2), matplotlib (3.4.3), and seaborn (0.11.2) for data handling and visualization.

\subsection{Cohort Selection}

In this study, we utilized the diabetes dataset \cite{UCI-Diabetes}, providing a depiction of clinical care across 130 US hospitals and integrated delivery networks spanning from 1999 to 2008. The primary classification objective involves predicting whether a patient is likely to be readmitted within 30 days. 
\textbf{Dataset Features:} Comprising data from 101,766 patients, the dataset encompasses 50 attributes, including 10 numerical, 7 binary, and 33 categorical features. The attributes encounter\_id and patient\_nbr are excluded from the learning tasks as they serve as patient identifiers. Additionally, attributes such as weight, payer\_code, and medical\_specialty are excluded due to containing at least 40\% missing values. Furthermore, we eliminate missing values in the race, diag\_1, diag\_2, and diag\_3 columns. The class label "readmitted" initially contains 54,864 rows with the designation "no record of readmission"; consequently, these rows are excluded as well. Following these data processing steps, the refined dataset consists of 45,715 records.\textbf{ Class Attribute:} The class attribute is the "readmitted" attribute, denoted as '< 30' and '> 30,' indicating whether a patient is likely to be readmitted within 30 days. The positive class is represented by "< 30." The dataset exhibits an imbalance with an Imbalance Ratio (IR) of 1:3:13 for the positive class to negative classes.

\section{Results}
\subsection{Quantifying overall model performance  }

Our study assesses ML models on a diabetes dataset, analyzing impacts by race, gender, and age. We start with model predictions, evaluating accuracy through metrics like accuracy, precision, recall, and F1 score. We then proceed to a fairness analysis using metrics like GSR, PPR, PPGR, FDR, FPR, FOR, and FNR. Additionally, we examine sub-group accuracy to ensure equitable performance. Through this process, we iteratively refine our models to enhance both accuracy and fairness, aiming to improve ML applications in healthcare, with a focus on equitable diabetes care.

\begin{table}
  \caption{Comparative Analysis of Machine Learning Model Performance}
  \label{model-performance-comparison}
  \centering
    \resizebox{\columnwidth}{!}{%
  \begin{tabular}{lllll}
    \toprule
    \textbf{Model name} & \textbf{Precision} & \textbf{Recall} & \textbf{F1} & \textbf{Accuracy} \\
    \midrule
   
    NB & 0.778 & 0.989 & 0.772 & 0.740 \\
     GLM & 0.797 & 0.989 & 0.830 & 0.809 \\
         GBM & \textbf{0.815} & \textbf{0.998} & \textbf{0.843} & \textbf{0.822} \\
             MLP& 0.792 & 0.998 & 0.839 & 0.819 \\

    \bottomrule
  \end{tabular}%
  }
\end{table}

Our evaluation of machine learning model performance, detailed in Table \ref{model-performance-comparison}, revealed GBM as the most comprehensive performer across the employed metrics (precision, recall, F1-score, and accuracy). Notably, GBM achieved a well-balanced F1-score of 84.3\%, demonstrating its proficiency in correctly identifying positive cases and capturing true positive cases. This balanced performance is further validated by GBM's accuracy of 82.2\%, indicating its ability to make accurate predictions. While other models exhibited strengths in specific areas, they presented trade-offs. For instance, MLP achieved a high precision of 79.2\%, suggesting its effectiveness in minimizing false positives. However, this strength came at the cost of a lower overall accuracy compared to GBM. Similarly, GLM and NB demonstrated competitive recall, indicating their ability to identify true positive cases. However, their lower accuracy (80.9\% for GLM and 74\% for NB) suggests potential limitations in correctly predicting positive cases overall.

\subsection{Quantifying model performance disparities across gender groups }
Table \ref{model-performance-gender} provides a gender-wise (male, female) comparison of classifiers, assessing their performance through key fairness metrics. This analysis aims to highlight each model's ability to achieve fairness and accuracy across genders, offering insights into their effectiveness in minimizing bias and maintaining equitable predictions.

\begin{table}[ht!]
  \caption{Model Performance Metrics Across Gender Categories}
  \label{model-performance-gender}
  \centering
    \resizebox{\columnwidth}{!}{%
  \begin{tabular}{lllllllll}
    \toprule
    \multicolumn{2}{c}{Group} & \multicolumn{7}{c}{Metrics} \\
    \cmidrule(r){1-2} \cmidrule(r){3-9}
    Gender & Model & GSR$\uparrow$ & PPR$\uparrow$ & PPGR$\uparrow$ & FDR$\downarrow$ & FPR$\downarrow$ & FOR$\downarrow$ & FNR$\downarrow$ \\
    \midrule
    \multirow{4}{*}{Female}  & NB  & 0.54 & 0.56 & 0.37 & 0.18 & 0.13 & 0.31 & 0.40 \\
                              
                            & GBM & 0.54 & 0.54 & 0.37 & 0.07 & 0.05 & 0.25 & 0.32 \\
                            & GLM & 0.55 & 0.54 & 0.39 & 0.09 & 0.07 & 0.24 & 0.29 \\
                              & MLP& 0.54 & 0.53 & 0.37 & 0.08 & 0.06 & 0.25 & 0.31 \\
                           
    \midrule
    \multirow{4}{*}{Male}   & NB  & 0.46 & 0.44 & 0.35 & 0.18 & 0.12 & 0.31 & 0.42 \\
                            & GBM & 0.46 & 0.46 & 0.37 & 0.06 & 0.05 & 0.23 & 0.30 \\
                            & GLM & 0.45 & 0.46 & 0.40 & 0.10 & 0.08 & 0.24 & 0.29 \\
                             & MLP& 0.46 & 0.47 & 0.39 & 0.10 & 0.07 & 0.23 & 0.29 \\

    \bottomrule
  \end{tabular}%
  }
\end{table}

 All models reflected the dataset's inherent gender balance (GSR close to 0.5 for both genders), ensuring the models were not skewed by disproportionate representation. However, there were some variations in how models treated the two groups for positive predictions. The NB model exhibited a bias towards females, with a higher PPR for females (56\%) compared to males (44\%). This could be due to bias in the model or reflect real trends in the data. The MLP model maintained a balanced approach with a slightly higher PPGR for males (1.05) than females, suggesting the model considers gender in positive predictions. 
 
 To evaluate overall accuracy and minimize errors, metrics like FDR and FPR were analyzed. The GBM model achieved the lowest error rates (FDR: 7\% females, 6\% males; FPR: 5\% for both genders), demonstrating its accuracy in minimizing false positives and correctly identifying negative cases. On the other hand, metrics like FOR and FNR measured how well models identified true positives. The GLM performed better for females with a lower FOR (24\%) and FNR (29\%), suggesting its efficiency in identifying true positives for females. However, the NB model indicated it might be missing important information when making predictions about females. (highest FOR: 31\% females\; highest FNR: 40\% females),

\subsection{Quantifying model performance disparities across race groups }

Table \ref{model-performance-race} provides a race-
wise comparison of classifiers, assessing their performance through different fairness metrics. This analysis is crucial for understanding each model's ability to deliver fairness and accuracy across different racial groups (agrican-american, caucasian, hispanic, asian), highlighting their effectiveness in minimizing bias and ensuring equitable predictions.

\begin{table}[ht!]
  \caption{Model Performance Metrics Across Racial Groups}
  \label{model-performance-race}
  \centering
  \resizebox{\columnwidth}{!}{%
  \begin{tabular}{lllllllll}
    \toprule
    \multicolumn{2}{c}{Group} & \multicolumn{7}{c}{Metrics} \\
    \cmidrule(r){1-2} \cmidrule(r){3-9}
    Race & Model & GSR$\uparrow$ & PPR$\uparrow$ & PPGR$\uparrow$ & FDR$\downarrow$ & FPR$\downarrow$ & FOR$\downarrow$ & FNR$\downarrow$ \\
    \midrule
    \multirow{4}{*}{African American}  & NB  & 0.19 & 0.20 & 0.36 & 0.20 & 0.14 & 0.31 & 0.40 \\ 
                                      & GBM & 0.19 & 0.19 & 0.37 & 0.08 & 0.05 & 0.23 & 0.30 \\
                                      & GLM & 0.19 & 0.19 & 0.39 & 0.11 & 0.08 & 0.24 & 0.29 \\
                                      & MLP & 0.19 & 0.20 & 0.38 & 0.11 & 0.08 & 0.23 & 0.29 \\
    \midrule
    \multirow{4}{*}{Asian}  & NB  & 0.01 & 0.01 & 0.36 & 0.00 & 0.00 & 0.48 & 0.46 \\
                                      & GBM & 0.01 & 0.01 & 0.58 & 0.00 & 0.00 & 0.20 & 0.12 \\
                                      & GLM & 0.01 & 0.01 & 0.41 & 0.06 & 0.05 & 0.27 & 0.29 \\
                                      & MLP & 0.01 & 0.01 & 0.56 & 0.00 & 0.00 & 0.25 & 0.17 \\
    \midrule
    \multirow{4}{*}{Caucasian}  & NB  & 0.77 & 0.77 & 0.36 & 0.18 & 0.13 & 0.31 & 0.41 \\
                                      & GBM & 0.77 & 0.76 & 0.36 & 0.07 & 0.05 & 0.24 & 0.32 \\
                                      & GLM & 0.77 & 0.77 & 0.39 & 0.09 & 0.07 & 0.24 & 0.29 \\
                                      & MLP & 0.77 & 0.76 & 0.37 & 0.08 & 0.06 & 0.24 & 0.30 \\
    \midrule
    \multirow{4}{*}{Hispanic}  & NB  & 0.02 & 0.02 & 0.34 & 0.15 & 0.11 & 0.39 & 0.47 \\
                                      & GBM & 0.02 & 0.02 & 0.39 & 0.04 & 0.04 & 0.28 & 0.32 \\
                                      & GLM & 0.02 & 0.02 & 0.39 & 0.10 & 0.08 & 0.31 & 0.35 \\
                                      & MLP & 0.02 & 0.02 & 0.38 & 0.13 & 0.11 & 0.35 & 0.39 \\
    \midrule
    \multirow{4}{*}{Other}  & NB  & 0.01 & 0.01 & 0.45 & 0.09 & 0.11 & 0.40 & 0.35 \\
                                      & GBM & 0.01 & 0.02 & 0.50 & 0.08 & 0.11 & 0.34 & 0.27 \\
                                      & GLM & 0.01 & 0.01 & 0.41 & 0.16 & 0.13 & 0.30 & 0.34 \\
                                      & MLP & 0.01 & 0.02 & 0.50 & 0.08 & 0.11 & 0.34 & 0.27 \\
    \bottomrule
  \end{tabular}%
  }
\end{table}

While all models reflected the dataset's inherent racial balance, some variations emerged in positive predictions. The NB model exhibited a tendency towards slightly higher positive predictions for certain groups, such as African Americans (20\% PPR), compared to other groups like Caucasians (19\% PPR in both MLP and GBM models). This suggests a potential bias in NB's positive prediction behavior. Conversely, the GBM model demonstrated a more balanced approach, maintaining similar PPGR across racial groups.

 GBM was a better performer in comparison to other models. It achieved the lowest error rates, indicated by a low FDR of 8\% for African Americans and 7\% for Caucasians. Additionally, GBM effectively identified negative cases with a low FPR of 5\% for both racial groups. In contrast, the NB model exhibited higher error rates, particularly for African Americans (FDR of 20\%), suggesting a higher proportion of incorrect positive predictions. This trend continued with identifying true positives. GBM performed well for African Americans with a lower FOR of 23\% and FNR of 30\%. Conversely, the NB model exhibited the most difficulty in accurately classifying true positives for African Americans, evident from its higher FOR (31\%) and FNR (40\%).

\subsection{Quantifying model performance disparities across age groups }

Table \ref{model-performance-age} summarizes the age-wise performance of classifiers, evaluating their effectiveness through different fairness metrics. This analysis seeks to shed light on how well each model manages to ensure accuracy and fairness across age groups (<40 and 40-99), pinpointing their capacity to avoid bias and promote equitable outcomes.

\begin{table}[ht!]
  \caption{Model Performance Metrics Across Different Age Groups}
  \label{model-performance-age}
  \centering
    \resizebox{\columnwidth}{!}{%
  \begin{tabular}{lllllllll}
    \toprule
    \multicolumn{2}{c}{Group} & \multicolumn{7}{c}{Metrics} \\
    \cmidrule(r){1-2} \cmidrule(r){3-9}
    Age Group & Model & GSR & PPR & PPGR & FDR & FPR & FOR & FNR \\
    \midrule
    \multirow{4}{*}{<40}   & MLP  & 0.11 & 0    & 0    & undefined & 0    & 0.27 & 1    \\
                           & GBM & 0.11 & 0.01 & 0.04 & 0.54      & 0.03 & 0.26 & 0.93 \\
                           & GLM & 0.11 & 0.05 & 0.19 & 0.65      & 0.17 & 0.28 & 0.78 \\
                           & NB  & 0.11 & 0.10 & 0.32 & 0.69      & 0.30 & 0.25 & 0.62 \\
    \midrule
    \multirow{4}{*}{40-59} & MLP  & 0.27 & 0.27 & 0.38 & 0.18 & 0.12 & 0.20 & 0.29 \\
                           & GBM & 0.27 & 0.22 & 0.29 & 0.07 & 0.04 & 0.24 & 0.38 \\
                           & GLM & 0.28 & 0.19 & 0.27 & 0.04 & 0.02 & 0.23 & 0.40 \\
                           & NB  & 0.27 & 0.21 & 0.27 & 0.16 & 0.08 & 0.29 & 0.47 \\
    \midrule
    \multirow{4}{*}{60-79} & MLP  & 0.35 & 0.61 & 0.66 & 0.06 & 0.13 & 0.21 & 0.11 \\
                           & GBM & 0.35 & 0.64 & 0.66 & 0.07 & 0.14 & 0.21 & 0.11 \\
                           & GLM & 0.35 & 0.60 & 0.66 & 0.07 & 0.14 & 0.23 & 0.11 \\
                           & NB  & 0.35 & 0.52 & 0.53 & 0.09 & 0.16 & 0.44 & 0.30 \\
    \midrule
    \multirow{4}{*}{80-99} & MLP  & 0.26 & 0.11 & 0.16 & 0.00 & 0.00 & 0.27 & 0.58 \\
                           & GBM & 0.26 & 0.13 & 0.19 & 0.03 & 0.01 & 0.26 & 0.54 \\
                           & GLM & 0.26 & 0.16 & 0.24 & 0.07 & 0.03 & 0.24 & 0.45 \\
                           & NB  & 0.26 & 0.17 & 0.23 & 0.18 & 0.07 & 0.26 & 0.52 \\

    \bottomrule
  \end{tabular}%
  }
\end{table}

All models reflected a balanced representation of age groups in the data (similar GSR across models for each age group). However, variations emerged in PPR).The NB model tended to under-predict positive outcomes for younger individuals (<40 years old, with a PPR of 10\%), while GBM tended to predict more positives for older adults (64\% PPR in the 60-79 age group). This trend is further supported by PPGR. Both MLP and GBM showed high PPGR values for the 60-79 age group (around 66\%), suggesting a bias towards positive predictions for older individuals.

Accuracy varied across models and age groups. MLP did not make any positive predictions for the youngest age group, resulting in an undefined FDR. Conversely, the GLM achieved a low FDR of 4\% in the 40-59 age group, indicating high accuracy in positive predictions for middle-aged adults. Interestingly, MLP showed perfect accuracy in identifying negative cases for the oldest age group (80-99 years old) with a FPR of 0

However, these findings also reveal potential shortcomings. The high FOR of 44\% for NB in the 60-79 age group suggests it misses many true positive cases among older adults. Similarly, the high FNR of 58\% for MLP in the 80-99 age group indicates a tendency to overlook true positives in the oldest individuals.

\begin{table}[ht!]
  \caption{Best Performing Models Across Categories}
  \label{best-performance-models}
  \centering
  \begin{tabular}{llll}
    \toprule
    Category & Group & Better Performance Model & PPGR \\
    \midrule
    \multirow{2}{*}{Gender} & Female & GLM & 39\% \\
                            & Male   & GBM & 40\% \\
    \midrule
    \multirow{5}{*}{Race}   & African American & GLM & 39\% \\
                            & Asian            & GBM & 58\% \\
                            & Caucasian        & GLM & 39\% \\
                            & Hispanic         & GBM, GLM & 39\% \\
                            & Other            & MLP, GBM & 50\% \\
    \midrule
    \multirow{4}{*}{Age}    & 40-59            & MLP & 38\% \\
                            & 60-79            & MLP, GBM, GLM & 66\% \\
                            & 80-99            & GLM & 24\% \\
                            & <40              & NB & 32\% \\
    \bottomrule
  \end{tabular}
\end{table}

\textbf{Summary:}
Table VI presents an overview of the highest-performing models across various categories under study. In our analysis, the GLM emerged as a top performer across several groups, achieving a PPGR of 39\% in both gender categories (female and male) and in the racial categories of African American, Caucasian, and Hispanic. Furthermore, GLM demonstrated better performance in comparison to other model, in the age groups of 60-79 and 80-99, with PPGRs of 66\% and 24\%, respectively, highlighting its broad applicability and effectiveness across diverse demographic segments. The GBM, on the other hand, showed better performance in the race categories of Asian, Hispanic, and Other, with PPGRs of 58\%, 39\%, and 50\%, respectively, and also excelled in the 60-79 age group with a PPGR of 66\%. MLP was identified as the top model in the 'Other' race category with a 50\% PPGR and in the age categories of 40-59 and 60-79, with PPGRs of 38\% and 66\%, respectively, underscoring its effectiveness in handling complex patterns in diabetes data. NB stood out in the age group of under 40 with a PPGR of 32\%, indicating its potential in early diabetes detection among younger populations. 

African American and Hispanic groups are affected by higher FDR and FPR rates across several models, resulting in a higher likelihood of receiving false-positive diagnoses. Notably, the NB model demonstrates a marked increase in FDR and FPR for African Americans, raising concerns about the accuracy of positive predictions for this demographic. While the Hispanic group encounters somewhat lower FDR overall, models like MLP still exhibit relatively higher FDR and FPR levels for this group, which may contribute to biased treatment outcomes.

The GBM model excels in delivering fair outcomes across diverse demographic segments. It exhibits consistently low rates of FDR and FPR for most racial and gender categories, alongside maintaining competitive predictive power as measured by the PPGR. Notably, when focusing on age, GBM, along with MLP and GLM, presents a well-balanced approach. This is particularly evident within the 60-79 age demographic, where it achieves a high PPGR, underscoring its efficiency and fairness in forecasting diabetes-related results for a wide range of patients.

Therefore, GBM emerges as the most equitable model overall, demonstrating a consistent ability to minimize biases and ensure fair treatment across different racial, gender, and age groups. This analysis underscores the importance of choosing and refining ML models that not only perform accurately but also uphold the principles of fairness and equity in healthcare predictions and treatments.
\section{Discussion}

Our  study emphasizes the crucial role of ML in advancing the diagnosis, treatment, and management of diabetes, a significant global health challenge. Through our evaluation, GBM emerged as the most effective model across a variety of performance metrics, including precision, recall, F1-score, and accuracy. GBM's balanced F1-score and commendable accuracy rate underscore its proficiency in accurately identifying positive cases and making precise predictions across different demographics.

This standout performance of GBM, however, is part of a larger narrative that includes the differential efficacy of models such as GLM, GBM, and MLP across diverse groups. For example, GLM shows enhanced performance for women and African Americans, while GBM excels with Asians and Hispanics. Such variations in model performance underscore the absence of a one-size-fits-all solution and highlight the critical need for selecting and tailoring models based on specific demographic groups to ensure accurate and equitable predictions.

The PPGR, a key finding of our study, further stresses the imperative for fairness in ML predictions. The variance in PPGR across models and demographics, especially with the lowest PPGR observed for the youngest age group by NB, suggests areas for improvement in equitable disease identification across age groups.

By identifying which models yield more accurate predictions for specific demographic groups, healthcare providers can better customize care, moving closer to the ideal of personalized medicine. The improved performance of GLM for certain groups, for instance, suggests its potential for developing targeted screening tools, thereby enhancing early diagnosis and treatment effectiveness. Similarly, the effectiveness of GBM with other groups could inform the creation of intervention programs that are more likely to engage patients and improve outcomes.

Moreover, the general superiority of GBM across most metrics and groups, particularly in gender and racial categorizations, affirms its robustness for accurate and fair classification. This highlights the importance of model selection and tuning to avoid bias and ensure equitable treatment, significantly contributing to the advancement of personalized medicine and tailoring interventions to individual patient profiles for improved healthcare outcomes.

The gender-based analysis of ML models for diagnosing diabetes highlights significant progress in achieving both technical accuracy and clinical fairness, particularly through the use of GBM. GBM's excellent performance, characterized by low error rates and well-balanced fairness metrics for both genders, demonstrates its effectiveness in providing accurate and unbiased predictions critical for the diagnosis and treatment of diabetes. This is in contrast to the NB model, which exhibited higher error rates, indicating challenges in maintaining both accuracy and fairness and underscoring the vital importance of precise model selection and calibration.

For women, GBM's outstanding performance is marked by the lowest FDR and FPR, showcasing its technical superiority in minimizing incorrect positive diagnoses. This high level of precision has direct clinical implications, leading to more dependable screenings and diagnoses for women, which is crucial for reducing the risk of misdiagnosis and ensuring fair and accurate care. GBM's balanced PPR and PPGR further highlight its effectiveness in providing equitable care, contributing to the reduction of gender disparities in healthcare outcomes. The contrast with the higher error rates of the NB model for women underscores the ongoing challenges in achieving fairness, highlighting the need for meticulous model selection and tuning in healthcare applications focused on women.

For men, GBM again demonstrates superior performance through minimal FDR and FPR, underscoring its ability to accurately classify male patients and reduce false positives. Such accuracy is essential for ensuring that men receive appropriate and timely care for diabetes, which has a significant impact on clinical outcomes by preventing both over-treatment and under-treatment. GBM's notable balance in PPGR and PPR metrics for males illustrates its capacity to maintain fairness, ensuring that diagnostic and treatment decisions are free from gender bias. Meanwhile, the increased likelihood of misclassification by the NB model in the male category indicates potential biases, emphasizing the importance of utilizing models like GBM that consistently exhibit high levels of precision and fairness across genders.

Our race-based examination in the context of diabetes care with ML models illuminates the critical importance of selecting models that cater well to specific racial groups, thereby enhancing both accuracy and fairness in healthcare outcomes. GLM and GBM stand out for their suitability across different racial demographics, illustrating how strategic model choice can contribute to reducing racial disparities in diabetes care. This approach underscores a movement toward a healthcare system that is equitable, by leveraging models that offer precision and fairness tailored to diverse racial groups.

For African Americans, the adept performance of GLM, indicated by its ability to effectively capture the unique healthcare needs of this demographic, underscores the model's potential in facilitating targeted and efficient diabetes interventions. This tailored approach can significantly improve clinical outcomes through early detection and management strategies that are sensitive to the distinct risk profiles prevalent within the African American community.

The Asian demographic benefits from the high predictive accuracy of GBM, highlighting the model's capability in addressing the specific healthcare requirements of Asians. Such accuracy is vital for creating diabetes care programs that are both culturally attuned and medically precise, thereby fostering better patient engagement and adherence to treatment.

In the case of Caucasians, GLM's adaptability, demonstrated by its effective performance, plays a crucial role in fine-tuning diabetes care to align with the epidemiological trends and healthcare needs of Caucasian patients. This adaptability is key to advancing personalized healthcare, enhancing the relevance and effectiveness of diabetes interventions.

The Hispanic demographic showcases the dual effectiveness of both GBM and GLM, each achieving notable performance. This reflects the diversity within the Hispanic community and signals the need for a healthcare approach that blends data-driven insights with cultural sensitivity to optimize diabetes care outcomes.

Lastly, the category labeled "Other" emphasizes the significance of inclusivity and representation in healthcare data. The substantial performance of both MLP and GBM models for this group points to the necessity of ensuring ML models are trained on diverse datasets. This is crucial for minimizing biases and improving the models' capacity to deliver equitable and accurate predictions across all racial groups, further advocating for the use of data to foster a more inclusive healthcare environment

The age-based analysis of ML models for diabetes care reveals significant insights into how different models can optimize healthcare delivery across various age groups, demonstrating the importance of selecting age-appropriate models to enhance accuracy and fairness in treatment. This analysis brings to light models like MLP, GLM, GBM, and NB for their distinct effectiveness across different age brackets.

For the younger population (under 40), the NB model stands out for its efficacy, despite higher FDR and FPR compared to other age groups. This effectiveness in predicting diabetes outcomes for younger individuals accentuates the challenge of maintaining fairness and underscores the need for model refinement to minimize bias and ensure that younger individuals receive equitable healthcare attention.

In the middle-aged group (40-59), MLP models shine with their high PPGR, balanced by relatively even FDR and FPR. These metrics suggest that MLP models offer not only effectiveness but also fairness in identifying diabetes-related outcomes among middle-aged individuals, highlighting the importance of bias-free models that accurately reflect diabetes prevalence and ensure inclusive, targeted interventions.

For older adults (60-79), the combined performance of MLP, GBM, and GLM models, as indicated by their PPGR and fairness metrics (FDR and FPR), showcases their ability to achieve both high accuracy and fairness. This dual achievement is essential for minimizing the risks of over- or under-diagnosing diabetes in older adults, suggesting that these models can significantly contribute to fair and effective diabetes management strategies for an age group that is particularly vulnerable to diabetes complications.

The eldest cohort (80-99) sees GLM as particularly effective, marked by high accuracy and a low FNR, complemented by favorable fairness metrics (lower FDR and FPR). This indicates that GLM provides a fair assessment of diabetes risk among the elderly, crucial for avoiding unnecessary interventions or missing treatment opportunities. This balance of accuracy and fairness is vital for supporting the healthcare needs of the elderly, ensuring interventions are both efficacious and equitable.

Overall, the age category conclusions reveal MLP and GLM as the most effective models in two out of four age groups, specifically for the 40-59 and 60-79 age brackets. GBM and NB each excel in one age group, with GBM suited for individuals in the 60-79 age range and NB for those under 40. This differentiation underscores the nuanced nature of model selection based on age, pushing towards a healthcare paradigm that emphasizes personalized care and mitigates age-related disparities in diabetes outcomes.

The broader implications of these findings underscore the practical impact of fairness analysis in ML models, ranging from promoting equity and equitable decision-making, mitigating bias, ensuring legal compliance, enhancing trust, and fostering positive societal outcomes. It highlights the critical role of ethical considerations in the development and deployment of AI technologies, reinforcing the need for careful model selection and customization to meet the diverse needs of patients across all age groups.

\paragraph{Limitations}
Our study offers a detailed examination but has some limitations worth noting. First, the reliance on conventional ML models may not capture the full complexity of diabetes-related health outcomes across different demographic groups, potentially oversimplifying the nuanced interplay of biological, environmental, and social determinants of health. Additionally, the fairness metrics used, though insightful, may not fully account for all dimensions of bias and equity, particularly in the context of intersectionality where multiple demographic factors overlap. The generalizability of findings is also limited by the specific dataset and metrics employed, which may not be representative of broader populations or encompass all relevant outcomes and predictors of diabetes.

\paragraph{Future directions}
Future work should aim to address these limitations through the incorporation of more advanced techniques, such as advanced deep learning architectures and ensemble methods, which may offer a more nuanced understanding of the patterns and predictors of diabetes across diverse populations. Further exploration into comprehensive fairness metrics and methodologies is also critical, including the development of models that can account for and mitigate multiple, intersecting biases. Expanding datasets to include a wider range of demographic, socio-economic, and health-related variables would enhance the robustness and applicability of findings. Lastly, integrating patient and clinician perspectives could offer invaluable insights into the practicality, acceptability, and ethical considerations of implementing ML models in clinical settings, ensuring that future advancements not only push the boundaries of technical feasibility but also align with the principles of patient-centered and equitable care.

\section{Conclusion}
Our study reveals GBM as a standout in applying ML to diabetes care. Notably, GBM demonstrates equitable performance across demographic lines.  Both GBM and GLM excel in delivering accurate and fair predictions, leading to improved outcomes for patients of different genders, races, and ages. This progress moves us closer to reducing disparities in diabetes care. However, challenges remain, particularly for African American and Hispanic groups who experience higher false positive rates. This highlights the importance of precise model selection. The success of GBM and GLM underscores the potential for more personalized and impactful healthcare interventions. Achieving this necessitates an interdisciplinary healthcare approach. Merging ML with broader medical expertise can deepen our understanding of diabetes and ensure a healthcare system that is both fair and responsive to all patients' needs. Our findings advocate for refining these technological tools to ensure they contribute positively to healthcare.  Here, fairness and personalization should be emphasized to guide future research towards a more equitable healthcare model for diabetes.

\bibliographystyle{IEEEtran} 
\bibliography{ref}

\begin{thebibliography}{10}
\providecommand{\url}[1]{#1}
\csname url@samestyle\endcsname
\providecommand{\newblock}{\relax}
\providecommand{\bibinfo}[2]{#2}
\providecommand{\BIBentrySTDinterwordspacing}{\spaceskip=0pt\relax}
\providecommand{\BIBentryALTinterwordstretchfactor}{4}
\providecommand{\BIBentryALTinterwordspacing}{\spaceskip=\fontdimen2\font plus
\BIBentryALTinterwordstretchfactor\fontdimen3\font minus \fontdimen4\font\relax}
\providecommand{\BIBforeignlanguage}[2]{{%
\expandafter\ifx\csname l@#1\endcsname\relax
\typeout{** WARNING: IEEEtran.bst: No hyphenation pattern has been}%
\typeout{** loaded for the language `#1'. Using the pattern for}%
\typeout{** the default language instead.}%
\else
\language=\csname l@#1\endcsname
\fi
#2}}
\providecommand{\BIBdecl}{\relax}
\BIBdecl

\bibitem{Braveman2014}
\BIBentryALTinterwordspacing
P.~Braveman and L.~Gottlieb, ``The social determinants of health: It’s time to consider the causes of the causes,'' \emph{Public Health Reports}, vol. 129, no. 1\_suppl2, p. 19–31, Jan. 2014. [Online]. Available: \url{http://dx.doi.org/10.1177/00333549141291S206}
\BIBentrySTDinterwordspacing

\bibitem{nihRootCauses}
``{T}he {R}oot {C}auses of {H}ealth {I}nequity --- ncbi.nlm.nih.gov,'' \url{https://www.ncbi.nlm.nih.gov/books/NBK425845/}, [Accessed 04-03-2024].

\bibitem{Williams2009-sz}
D.~R. Williams and S.~A. Mohammed, ``\BIBforeignlanguage{en}{Discrimination and racial disparities in health: evidence and needed research},'' \emph{\BIBforeignlanguage{en}{J. Behav. Med.}}, vol.~32, no.~1, pp. 20--47, Feb. 2009.

\bibitem{raza2023fairness}
S.~Raza, P.~O. Pour, and S.~R. Bashir, ``Fairness in machine learning meets with equity in healthcare,'' in \emph{Proceedings of the AAAI Symposium Series}, vol.~1, no.~1, 2023, pp. 149--153.

\bibitem{raza2023auditing}
S.~Raza and S.~R. Bashir, ``Auditing icu readmission rates in an clinical database: An analysis of risk factors and clinical outcomes,'' in \emph{2023 IEEE 11th International Conference on Healthcare Informatics (ICHI)}.\hskip 1em plus 0.5em minus 0.4em\relax IEEE, 2023, pp. 722--726.

\bibitem{Williams2013}
\BIBentryALTinterwordspacing
D.~R. Williams and S.~A. Mohammed, ``Racism and health i: Pathways and scientific evidence,'' \emph{American Behavioral Scientist}, vol.~57, no.~8, p. 1152–1173, May 2013. [Online]. Available: \url{http://dx.doi.org/10.1177/0002764213487340}
\BIBentrySTDinterwordspacing

\bibitem{10.1145/3287560.3287598}
\BIBentryALTinterwordspacing
A.~D. Selbst, D.~Boyd, S.~A. Friedler, S.~Venkatasubramanian, and J.~Vertesi, ``Fairness and abstraction in sociotechnical systems,'' in \emph{Proceedings of the Conference on Fairness, Accountability, and Transparency}, ser. FAT* '19.\hskip 1em plus 0.5em minus 0.4em\relax New York, NY, USA: Association for Computing Machinery, 2019, p. 59–68. [Online]. Available: \url{https://doi.org/10.1145/3287560.3287598}
\BIBentrySTDinterwordspacing

\bibitem{Berendt2017}
\BIBentryALTinterwordspacing
B.~Berendt and S.~Preibusch, ``Toward accountable discrimination-aware data mining: The importance of keeping the human in the loop—and under the looking glass,'' \emph{Big Data}, vol.~5, no.~2, p. 135–152, Jun. 2017. [Online]. Available: \url{http://dx.doi.org/10.1089/big.2016.0055}
\BIBentrySTDinterwordspacing

\bibitem{Reddy2019}
\BIBentryALTinterwordspacing
S.~Reddy, S.~Allan, S.~Coghlan, and P.~Cooper, ``A governance model for the application of ai in health care,'' \emph{Journal of the American Medical Informatics Association}, vol.~27, no.~3, p. 491–497, Nov. 2019. [Online]. Available: \url{http://dx.doi.org/10.1093/jamia/ocz192}
\BIBentrySTDinterwordspacing

\bibitem{oreillyPracticalFairness}
{Aileen Nielsen}, ``{P}ractical {F}airness --- oreilly.com,'' \url{https://www.oreilly.com/library/view/practical-fairness/9781492075721/}, [Accessed 05-03-2024].

\bibitem{raza2023connecting}
S.~Raza, ``Connecting fairness in machine learning with public health equity,'' in \emph{2023 IEEE 11th International Conference on Healthcare Informatics (ICHI)}.\hskip 1em plus 0.5em minus 0.4em\relax IEEE, 2023, pp. 704--708.

\bibitem{Kamiran2011}
\BIBentryALTinterwordspacing
F.~Kamiran and T.~Calders, ``Data preprocessing techniques for classification without discrimination,'' \emph{Knowledge and Information Systems}, vol.~33, no.~1, p. 1–33, Dec. 2011. [Online]. Available: \url{http://dx.doi.org/10.1007/s10115-011-0463-8}
\BIBentrySTDinterwordspacing

\bibitem{Wexler_2019}
\BIBentryALTinterwordspacing
J.~Wexler, M.~Pushkarna, T.~Bolukbasi, M.~Wattenberg, F.~Viegas, and J.~Wilson, ``The what-if tool: Interactive probing of machine learning models,'' \emph{IEEE Transactions on Visualization and Computer Graphics}, p. 1–1, 2019. [Online]. Available: \url{http://dx.doi.org/10.1109/TVCG.2019.2934619}
\BIBentrySTDinterwordspacing

\bibitem{pleiss2017fairness}
G.~Pleiss, M.~Raghavan, F.~Wu, J.~Kleinberg, and K.~Q. Weinberger, ``On fairness and calibration,'' 2017.

\bibitem{10.1145/3278721.3278729}
\BIBentryALTinterwordspacing
L.~Dixon, J.~Li, J.~Sorensen, N.~Thain, and L.~Vasserman, ``Measuring and mitigating unintended bias in text classification,'' in \emph{Proceedings of the 2018 AAAI/ACM Conference on AI, Ethics, and Society}, ser. AIES '18.\hskip 1em plus 0.5em minus 0.4em\relax New York, NY, USA: Association for Computing Machinery, 2018, p. 67–73. [Online]. Available: \url{https://doi.org/10.1145/3278721.3278729}
\BIBentrySTDinterwordspacing

\bibitem{Holstein_2019}
\BIBentryALTinterwordspacing
K.~Holstein, J.~Wortman~Vaughan, H.~Daumé, M.~Dudik, and H.~Wallach, ``Improving fairness in machine learning systems: What do industry practitioners need?'' in \emph{Proceedings of the 2019 CHI Conference on Human Factors in Computing Systems}, ser. CHI ’19.\hskip 1em plus 0.5em minus 0.4em\relax ACM, May 2019. [Online]. Available: \url{http://dx.doi.org/10.1145/3290605.3300830}
\BIBentrySTDinterwordspacing

\bibitem{mayoclinicHyperglycemiaDiabetesHyperglycemia}
{Mayo Clinic}, ``{H}yperglycemia in diabetes-{H}yperglycemia in diabetes - {S}ymptoms \& causes - {M}ayo {C}linic --- mayoclinic.org,'' \url{https://www.mayoclinic.org/diseases-conditions/hyperglycemia/symptoms-causes/syc-20373631}, [Accessed 04-03-2024].

\bibitem{Strack2014}
\BIBentryALTinterwordspacing
B.~Strack, J.~P. DeShazo, C.~Gennings, J.~L. Olmo, S.~Ventura, K.~J. Cios, and J.~N. Clore, ``Impact of hba1c measurement on hospital readmission rates: Analysis of 70, 000 clinical database patient records,'' \emph{BioMed Research International}, vol. 2014, p. 1–11, 2014. [Online]. Available: \url{http://dx.doi.org/10.1155/2014/781670}
\BIBentrySTDinterwordspacing

\bibitem{pagano2022bias}
T.~P. Pagano, R.~B. Loureiro, F.~V.~N. Lisboa, G.~O.~R. Cruz, R.~M. Peixoto, G.~A. de~Sousa~Guimarães, L.~L. dos Santos, M.~M. Araujo, M.~Cruz, E.~L.~S. de~Oliveira, I.~Winkler, and E.~G.~S. Nascimento, ``Bias and unfairness in machine learning models: a systematic literature review,'' 2022.

\bibitem{Giovanola2022}
\BIBentryALTinterwordspacing
B.~Giovanola and S.~Tiribelli, ``Beyond bias and discrimination: redefining the ai ethics principle of fairness in healthcare machine-learning algorithms,'' \emph{AI \& SOCIETY}, vol.~38, no.~2, p. 549–563, May 2022. [Online]. Available: \url{http://dx.doi.org/10.1007/s00146-022-01455-6}
\BIBentrySTDinterwordspacing

\bibitem{Gohar_2023}
\BIBentryALTinterwordspacing
U.~Gohar and L.~Cheng, ``A survey on intersectional fairness in machine learning: Notions, mitigation, and challenges,'' in \emph{Proceedings of the Thirty-Second International Joint Conference on Artificial Intelligence}, ser. IJCAI-2023.\hskip 1em plus 0.5em minus 0.4em\relax International Joint Conferences on Artificial Intelligence Organization, Aug. 2023. [Online]. Available: \url{http://dx.doi.org/10.24963/ijcai.2023/742}
\BIBentrySTDinterwordspacing

\bibitem{chen2023comprehensive}
Z.~Chen, J.~M. Zhang, F.~Sarro, and M.~Harman, ``A comprehensive empirical study of bias mitigation methods for machine learning classifiers,'' 2023.

\bibitem{pessach2020algorithmic}
D.~Pessach and E.~Shmueli, ``Algorithmic fairness,'' 2020.

\bibitem{bdcc7010027}
\BIBentryALTinterwordspacing
T.~P. Pagano, R.~B. Loureiro, F.~V.~N. Lisboa, G.~O.~R. Cruz, R.~M. Peixoto, G.~A. d.~S. Guimarães, E.~L.~S. Oliveira, I.~Winkler, and E.~G.~S. Nascimento, ``Context-based patterns in machine learning bias and fairness metrics: A sensitive attributes-based approach,'' \emph{Big Data and Cognitive Computing}, vol.~7, no.~1, 2023. [Online]. Available: \url{https://www.mdpi.com/2504-2289/7/1/27}
\BIBentrySTDinterwordspacing

\bibitem{Wan2023}
\BIBentryALTinterwordspacing
M.~Wan, D.~Zha, N.~Liu, and N.~Zou, ``In-processing modeling techniques for machine learning fairness: A survey,'' \emph{ACM Transactions on Knowledge Discovery from Data}, vol.~17, no.~3, p. 1–27, Mar. 2023. [Online]. Available: \url{http://dx.doi.org/10.1145/3551390}
\BIBentrySTDinterwordspacing

\bibitem{Yang2023}
\BIBentryALTinterwordspacing
J.~Yang, A.~A.~S. Soltan, D.~W. Eyre, Y.~Yang, and D.~A. Clifton, ``An adversarial training framework for mitigating algorithmic biases in clinical machine learning,'' \emph{npj Digital Medicine}, vol.~6, no.~1, Mar. 2023. [Online]. Available: \url{http://dx.doi.org/10.1038/s41746-023-00805-y}
\BIBentrySTDinterwordspacing

\bibitem{Wang2023}
\BIBentryALTinterwordspacing
R.~Wang, P.~Chaudhari, and C.~Davatzikos, ``Bias in machine learning models can be significantly mitigated by careful training: Evidence from neuroimaging studies,'' \emph{Proceedings of the National Academy of Sciences}, vol. 120, no.~6, Jan. 2023. [Online]. Available: \url{http://dx.doi.org/10.1073/pnas.2211613120}
\BIBentrySTDinterwordspacing

\bibitem{wang2023fair}
Z.~Wang, Y.~Zhou, M.~Qiu, I.~Haque, L.~Brown, Y.~He, J.~Wang, D.~Lo, and W.~Zhang, ``Towards fair machine learning software: Understanding and addressing model bias through counterfactual thinking,'' 2023.

\bibitem{canadaSocialDeterminants}
{Public Health Agency of Canada}, ``{S}ocial determinants of health and health inequalities - {C}anada.ca --- canada.ca,'' \url{https://www.canada.ca/en/public-health/services/health-promotion/population-health/what-determines-health.html}, [Accessed 05-03-2024].

\bibitem{Mosca2011-vv}
L.~Mosca, E.~Barrett-Connor, and N.~K. Wenger, ``\BIBforeignlanguage{en}{Sex/gender differences in cardiovascular disease prevention: what a difference a decade makes},'' \emph{\BIBforeignlanguage{en}{Circulation}}, vol. 124, no.~19, pp. 2145--2154, Nov. 2011.

\bibitem{Hartley2004-vl}
D.~Hartley, ``\BIBforeignlanguage{en}{Rural health disparities, population health, and rural culture},'' \emph{\BIBforeignlanguage{en}{Am. J. Public Health}}, vol.~94, no.~10, pp. 1675--1678, Oct. 2004.

\bibitem{Koh2011-ru}
H.~K. Koh, G.~Graham, and S.~A. Glied, ``\BIBforeignlanguage{en}{Reducing racial and ethnic disparities: the action plan from the department of health and human services},'' \emph{\BIBforeignlanguage{en}{Health Aff. (Millwood)}}, vol.~30, no.~10, pp. 1822--1829, Oct. 2011.

\bibitem{Lopez2011-yf}
L.~L{\'o}pez, A.~R. Green, A.~Tan-McGrory, R.~King, and J.~R. Betancourt, ``\BIBforeignlanguage{en}{Bridging the digital divide in health care: the role of health information technology in addressing racial and ethnic disparities},'' \emph{\BIBforeignlanguage{en}{Jt. Comm. J. Qual. Patient Saf.}}, vol.~37, no.~10, pp. 437--445, Oct. 2011.

\bibitem{saleiro2019aequitas}
P.~Saleiro, B.~Kuester, L.~Hinkson, J.~London, A.~Stevens, A.~Anisfeld, K.~T. Rodolfa, and R.~Ghani, ``Aequitas: A bias and fairness audit toolkit,'' 2019.

\bibitem{UCI-Diabetes}
B.~Strack, J.~DeShazo, C.~Gennings, J.~Olmo, S.~Ventura, K.~Cios, and J.~Clore, ``{UCI} machine learning repository: Diabetes 130-{US} hospitals for years 1999–2008 dataset,'' \url{https://archive.ics.uci.edu/ml/datasets/diabetes+130-us+hospitals+for+years+1999-2008}, 2014.

\end{thebibliography}

\end{document}